%% file: samplepaper.tex
\documentclass[runningheads]{waica}
\usepackage[T1]{fontenc}
\usepackage{multicol}
\usepackage{amsmath}
\usepackage{amssymb}
\usepackage{booktabs}
\usepackage{multirow}
\usepackage{multicol}
\usepackage{graphicx}
\usepackage[bookmarks=true]{hyperref}
\usepackage{dblfloatfix} 
\usepackage[table]{xcolor}
\usepackage{float}
\usepackage{subfigure}
\usepackage{url}
\usepackage{enumitem}
\usepackage{wrapfig}
\usepackage{sidecap}
\begin{document}
\title{Act, Sense, Act: Learning Active Perception from Large-Scale Egocentric Human Data}
%

\author{\textbf{Jialiang Li}*, \textbf{Yi Qiao}*, \textbf{Yunhan Guo}, \textbf{Changwen Chen}, \textbf{Wenzhao Lian} \\
*Equal Contribution \\
}

\institute{School of Artificial Intelligence, Shanghai Jiao Tong University}
%
\authorrunning{Li et al.}
\titlerunning{Act-Sense-Act}
%
%

\setcounter{figure}{1}
\makeatletter
\let\@oldmaketitle\@maketitle
\renewcommand{\@maketitle}{\@oldmaketitle
  \begin{center}
    \includegraphics[width=\textwidth]{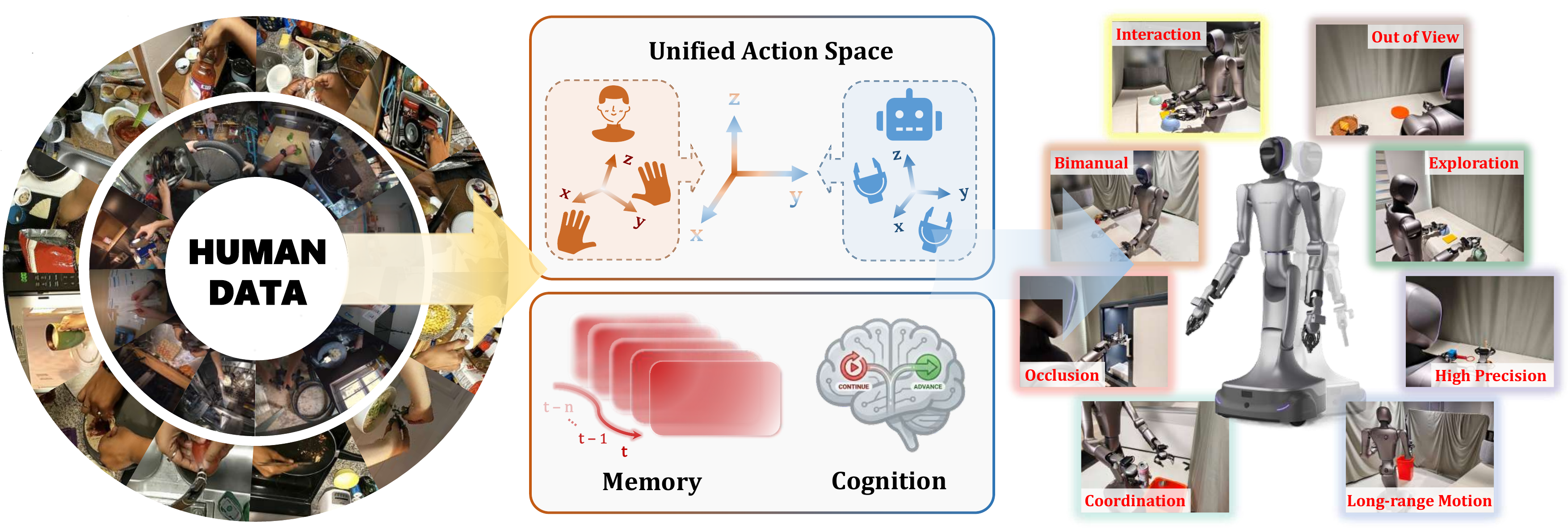}
    \label{fig:teaser}
  \end{center}
  \vspace{-15pt}
  \refstepcounter{figure}
  \small{\textbf{Fig.~\thefigure.}~\label{fig:teaser} By unifying large-scale egocentric human data and robot data within a shared action space, a wheel-based humanoid develops robust and adaptive active perception capabilities through memory-driven temporal and cognitive learning mechanisms. This establishes a tight \textbf{"Act, Sense, Act"} coupling between perception and action, which effectively addresses diverse active perception tasks.}
  \label{fig:teaser}
 \vspace{-15pt}
  \medskip}
\makeatother

\maketitle              
\begin{abstract}
Achieving generalizable manipulation in unconstrained environments requires the robot to proactively resolve information uncertainty, i.e., the capability of active perception. However, existing methods are often confined in limited types of sensing behaviors, restricting their applicability to complex environments. 
In this work, we formalize active perception as a history-dependent perception-action loop driven by information-seeking action and decision branching, providing a structured categorization of visual active perception paradigms. 
Building on this perspective, we introduce CoMe-VLA, a cognitive and memory-aware vision-language-action (VLA) framework that leverages large-scale human egocentric data to learn versatile exploration and manipulation priors. Our framework integrates a cognitive auxiliary head for autonomous sub-task transitions and a dual-track memory system to maintain consistent self and environmental awareness by fusing proprioceptive and visual temporal contexts. By aligning human and robot hand-eye coordination behaviors in a unified egocentric action space, we train the model progressively in three stages. Extensive experiments on a wheel-based humanoid have demonstrated strong robustness and adaptability of our proposed method across diverse long-horizon tasks spanning multiple active perception scenarios.View our project in \url{https://jern-li.github.io/asa/}

\end{abstract}

\input{sections/1_introduction}
\input{sections/2_related_work}

\input{sections/3_problem_definition}

\input{sections/4_data_collection}

\input{sections/5_method}

\input{sections/6_experiments}
\input{sections/7_conclusion}
\bibliographystyle{splncs04}
\bibliography{references}

\newpage
\input{sections/X_suppl}

\end{document}

%% file: sections/1_introduction.tex
\section{Introduction}

The significant progress in robotic manipulation in recent years has been largely driven by the rapid advancement of imitation learning (IL) \cite{act,chi2025diffusion} and foundation models \cite{kim2025openvla-oft,black2025pi_0_5}. These approaches have demonstrated promising performance on deterministic linear tasks in structured environments, where the observation-to-action mapping along the task execution is stationary. 
However, to deploy robots in complex unstructured environments, intentional gazing and exploration based on previous interaction history, rather than just reactive movement, are required.
For example, to retrieve a wrench buried in a cluttered toolbox, a robot may need to explore different viewpoints or manipulate surrounding objects to reveal occluded regions. The robot then continuously adapts its execution based on newly observed information, such as reaching towards the left if the wrench appears on the left, or towards the right if revealed on the right.
Such scenarios require a cognitive framework that resolves ambiguity with a coupled act-sense-act loop,
where perception provides evolving sensory information that triggers adaptive decisions, and decisions in turn guide actions to resolve perceptual ambiguity or adopt alternative manipulation strategies, to progress towards task success or generate new observations informing further decisions. 
This closed-loop process illustrates the core principle of \textbf{Active Perception}.

A few efforts \cite{nbp_observe-then-act,gaze_takizawa2025gaze,gaze_chuang2025look,activeumi,egomi,ViA} have been attempted to tackle the above mentioned perceptual passivity issue. 
However, these approaches treat active perception control (such as head/eye movement) merely as an extra action dimension, and optimize for immediate task completion. Consequently, these models are commonly confined to head movement-based viewpoint adjustment for linear task execution, failing to utilize body movement or interactive manipulation, such as opening a drawer, as strategic tools to reveal hidden information.

Considering the above problems, we frame active perception as a continuous \textbf{"Act, Sense, Act"} process. Specifically, we first advance the theoretical understanding of active perception, formalizing it as a history-dependent perception-action loop driven by information-seeking action and decision branching, while providing categorization of visual active perception paradigms; and then introduce CoMe-VLA, as illustrated in Figure \ref{fig:teaser}, a cognitive and memory-aware Vision-Language-Action (VLA) framework that leverages large-scale human egocentric data to enable robust manipulation across diverse active perception scenarios. Our approach is built upon two key insights. First, humans naturally perform complex active perception tasks every day (e.g., turning around to adjust viewpoints or opening a drawer to see its contents), inherently embedding rich priors of proactive exploration, adaptive decision-making, and manipulation. Second, effective long-horizon active perception requires agents to be both memory-aware and cognitively capable, enabling temporal context retention (e.g., "I have searched the left") and robust task progress estimation for adaptive execution (e.g., "target has been found and now I should fetch it").

Concretely, we first collect large-scale human egocentric data and align it with robot data in a unified egocentric action space, therefore bridging the cross-embodiment gap. Built upon a visual-language backbone, we design a cognitive auxiliary head for reliable sub-task completion detection and a dual-track memory system to aggregate historical visual and proprioceptive information. In addition, we employ a three-stage training strategy, ranging from cognitive pretraining on human data, full-model pretraining on human data, to full-model finetuning on robot data. Extensive experiments demonstrate the robustness and effectiveness of our approach across diverse long-horizon tasks spanning various visual active perception paradigms.

Our contributions are summarized as follows. 
\begin{itemize}[topsep=2pt]
    \item We formalize active perception as a history-dependent perception-action loop, which moves beyond reactive sensing by modeling perception as an intentional act driven by information-seeking action and decision branching, providing a systematic taxonomy for visual exploration.

    \item We introduce a method to distill exploratory priors from large-scale human egocentric datasets. By aligning human and robot coordination in a unified egocentric action space, we enable the transfer of human-like "act-sense-act" strategies to robot platforms.

    \item We propose CoMe-VLA, featuring a dual-track memory system and a cognitive auxiliary head. This enables the robot to maintain environmental awareness across long horizons and autonomously trigger sub-task transitions.
    
    \item Extensive experiments on a wheel-based humanoid demonstrate emergent active perception behaviors and robustness to dynamic perturbations. Our results show that human priors significantly reduce the requirement for robot-specific demonstrations while maintaining high success rates in complex, unconstrained environments.

\end{itemize}



%% file: sections/2_related_work.tex
\section{Related Work}

\subsection{Learning Manipulation from Human Data}
Human demonstrations provide a rich source of structured behavior and semantic information for robotic manipulation. Dedicated human demonstrations are commonly collected by recording humans performing tasks using external sensing setups \cite{dexcap,humanoid,emma}, or with specialized devices such as UMI \cite{umi,dexumi}, which provide precise motion trajectories but remain limited in task coverage and environmental diversity. To achieve greater scalability, recent works have turned to public egocentric datasets \cite{perrett2025hd,hoque2025egodex,kong2025aria} to learn robust representations, such as visual priors for scene understanding \cite{zhu2024vision}, functional affordances \cite{affordance_2handedafforder,affordance_agrl}, and high-level task semantics \cite{mavip,bharadhwaj2024towards}. However, regarding motion modeling, most methods primarily focus on aligning hand-centric action spaces to map visual observations to end-effector trajectories \cite{egovla,egomimic}. These approaches typically treat image observations as exogenous inputs, with limited consideration of how viewpoints are selected during manipulation. In this paper, we argue that the camera view is not a passive stream but an intentional result of eye-hand coordination. By incorporating head movements into action alignment,  our model learns the proactive coupling between where to look and how to act.

\subsection{Policy Learning with Active Perception} 
Early attempts to integrate active perception into robotic systems rely on heuristics, such as next-best-view \cite{nbv_guedon2023macarons,nbv_chen2024gennbv}, next-best-pose \cite{nbp_observe-then-act,nbp_graspview}, and gaze control \cite{gaze_takizawa2025gaze,gaze_chuang2025look}. However, these methods either decouple viewpoint selection from manipulation and require exhaustive iterations, or fail to proactively explore information out of sight, limiting their applicability to real-world dynamics. With the rise of end-to-end learning, recent imitation learning approaches have attempted to treat camera movement as a learnable action \cite{ViA,activeumi,egomi}. Yet, most remain limited to linear task execution, where sensing is treated as a head-pose extension rather than a strategic tool to resolve uncertainty. They often overlook the role of physical manipulation in overcoming occlusions or the action decision branching caused by different perceptual outcomes. In this paper, we advance beyond simple viewpoint adjustment by exploring the intrinsic mechanisms of active perception systematically, proposing a cognitive and memory-aware approach that ensures robust reasoning and adaptive execution in non-linear tasks.


%% file: sections/3_problem_definition.tex
\section{Problem Formulation}
\begin{figure}[htbp]
    \centering
    \begin{minipage}[c]{0.50\textwidth}
        \centering
        \includegraphics[width=\textwidth]{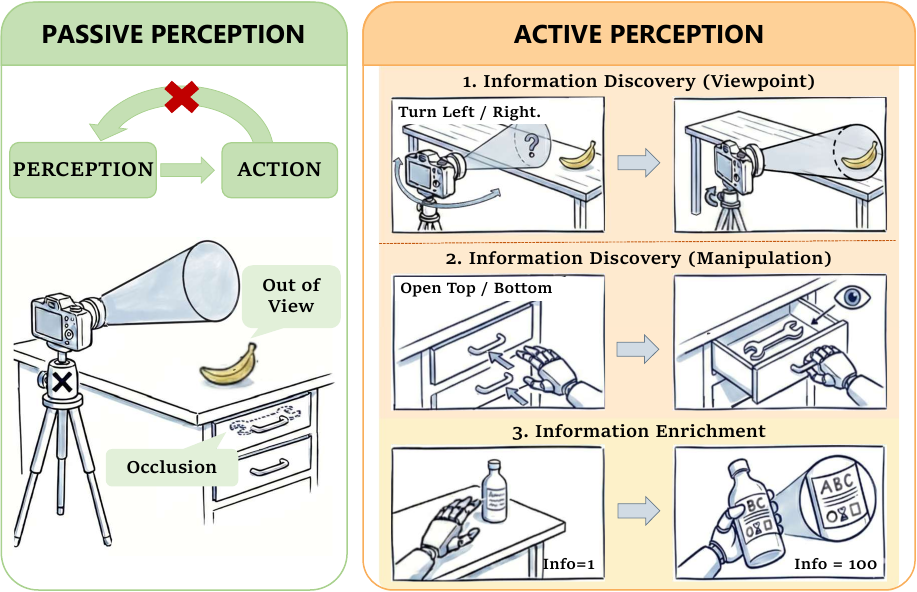}
    \end{minipage}
    \hfill 
    \begin{minipage}[c]{0.47\textwidth}
        \caption{Comparison between passive perception and active perception. \textbf{Left:} Passive perception, based on static observation with limited information when targets are out of view or occluded. \textbf{Right:} Active perception, which enables proactive actions to resolve task-relevant uncertainty, illustrated through three paradigms.}
        \label{fig_problem_formulation}
    \end{minipage}
    \vspace{-4pt}
\end{figure}

As illustrated in Figure \ref{fig_problem_formulation}, unlike passive perception that follows a linear perception-to-action flow, active perception establishes a closed-loop interaction between perception and action, where deliberate actions are executed to gather task-relevant information, and the resulting perceptual outcomes direct the branching of the subsequent actions. 


\subsection{Core Mechanisms for Active Perception}
\label{sec_core_mechanisms}

We describe active perception as a history-dependent perception-action loop centered on two core mechanisms:

\textbf{1) Information-Seeking Action.} Active perception works by taking purposeful actions to improve what the agent can see or sense. Instead of waiting for information to appear, the agent moves its viewpoint, adjusts its posture, or explores the scene to gather task-relevant information. 


\textbf{2) Decision Branching.} The agent’s behavior is not fixed, but dynamically branches based on what has been observed so far. Different perceptual contexts and accumulated histories naturally lead to distinct behavioral strategies.

Decision branching manifests through two forms:
\begin{itemize}[topsep=0pt]
    \item \textbf{Exploratory Branching.} The adaptation of information-seeking strategies based on perceptual evidence (e.g., gazing on the target when it is found, otherwise scanning).
    \item \textbf{Exploitative Branching.} The adaptation of manipulation strategies based on resolved perceptual states (e.g., executing a left-handed grasp if the target is positioned on the left, or a right-handed grasp if on the right).
\end{itemize}

\subsection{Paradigms of Visual Active Perception}
\label{sec_paradigm}
With respect to visual information, we categorize active perception into two distinct paradigms, as shown in the right plot of Figure \ref{fig_problem_formulation}:

\textbf{1) Information Discovery (ID).} Task-relevant visual information is initially absent, the agent employs proactive exploration strategies to intentionally bring the object of interest into its field of view. This paradigm is further divided based on the nature of the exploratory action:

\begin{itemize}[topsep=0pt]
    \item \textbf{Viewpoint Discovery (VD).} The agent changes its own viewpoint (e.g., head pan-tilt or base movement) to locate an object previously out of sight. 
    \item  \textbf{Manipulation Discovery (MD).} The agent physically interacts with the environment (e.g., opening a drawer) to reveal hidden information. 
\end{itemize}

\textbf{2) Information Enrichment (IE).} Task-relevant visual information is already present but insufficient for high-precision task completion, requiring actions to amplify resolution (e.g., grasping and bringing an object closer for detailed inspection and manipulation).

%% file: sections/4_data_collection.tex
\section{Data Collection}

\subsection{Large-Scale Egocentric Human Dataset}

Egocentric human data offers a direct window into how humans interact with the world under physical constraints, capturing perception, decision-making, and execution as they unfold in time. We therefore leverage publicly available human datasets, which are low-cost and efficient to acquire, to learn generalizable priors for active perception. Specifically, our data selection prioritizes: (1) \textbf{fine-grained annotations} for detailed action and state understanding; (2)  \textbf{synchronized hand and head/camera poses} to model eye-hand coordination; and (3) \textbf{scenarios reflecting active perception challenges}, such as cluttered environments with heavy occlusions, dynamic multi-object interactions, viewpoint adjustments to resolve ambiguity. Based on these criteria, we select CaptainCook4D \cite{peddi2024captaincook4d} and Ego-Exo4D \cite{grauman2024ego} as the primary data sources. 

\subsection{Egocentric Robot Teleoperation}

While in-the-wild human data provide generic priors, on-task robot data is essential for grounding these priors into specific environments and concrete embodiments. We thus collect robot-centric data using a Corenetic Monte02 humanoid \cite{corenetic} teleoperated via Meta Quest 3 \cite{meta}. This VR system maps operator head and hand movements to the robot’s head, arms,  grippers, and chassis in real-time, while streaming egocentric RGB observations back to the headset. More details are provided in the supplementary material.

\subsection{Human-Robot Data Alignment}

We formulate human–robot data alignment as a problem of structural isomorphism, representing both modalities in a unified, egocentric action space. This normalizes trajectories from both domains into a consistent, episode-relative reference frame, mitigating distribution shifts due to different embodiments and sensor configurations while preserving the intrinsic eye-hand coordination.

A key challenge in human data is the absence of a consistent global base frame across demonstrations (e.g., varying starting poses, environments, and recording setups). To address this, we define the episode base frame $\mathcal{B}$ as the initial frame of each trajectory, expressing subsequent poses relative to it. All poses are represented in a body-centric local frame $\mathcal{L}$, with the $\mathbf{x}$-axis pointing forward, $\mathbf{y}$-axis left, and  $\mathbf{z}$-axis upward. Under these conventions, both human and robot trajectories are mapped into the same egocentric, episode-relative space via a unified transformation:
\begin{gather}
T_i^B(t) = (T_B^W)^{-1} \cdot T_i^W(t) \cdot T_{i_{j} \to \mathcal{L}} , \quad
i \in \{H, L, R\},  \notag \\
\quad j \in \{\text{CaptainCook4D}, \text{Ego-Exo4D, Robot}\},
\end{gather}
where $H$ denotes the head, $L$ and $R$ the left and right wrists, $B$ the base frame,  and $W$ the world frame. 

To reconcile the morphological differences between human hands and parallel-jaw grippers, we map the high-dimensional hand configuration to a single gripper width, which is computed by averaging the distances from the thumb tip to other fingertips, abstracting complex finger articulations into a unified grasping signal. 

In human data, the egocentric perceptual reference is given by the recorded head pose. To mirror this representation on the robot, we aggregate chassis motion and head articulation into a composite head pose. During inference, this representation is decomposed via an inverse mapping to derive control commands for the chassis and head gimbal. Implementation details are provided in the supplementary material.

%% file: sections/5_method.tex
\section{Model Design}

\begin{figure}[t]
    \centering
    \includegraphics[width=\linewidth]{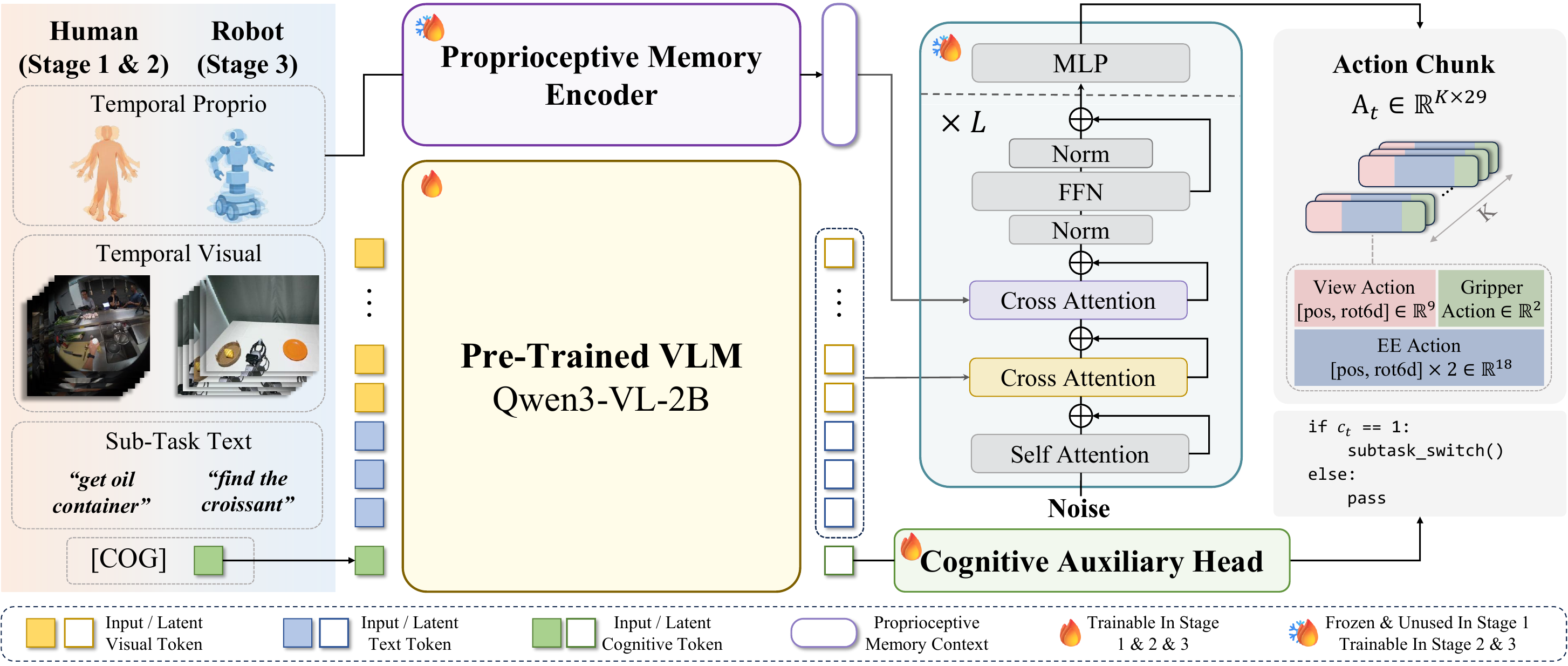}
    \setlength{\abovecaptionskip}{-8pt}
    \caption{\textbf{CoMe-VLA Overview.} CoMe-VLA integrates a pre-trained VLM (Qwen3-VL-2B \cite{bai2025qwen3vltechnicalreport}) with a transformer-based proprioceptive memory encoder to construct temporal visual-semantic and proprioceptive contexts, which are fed into a flow-matching action decoder to generate a 29-D action chunk. The VLM also outputs a cognitive latent token for the cognitive auxiliary head, which predicts a binary label for autonomous task transition. CoMe-VLA is trained in three stages using human and robot data. }
    \label{fig:model_architecture}
\end{figure}

As illustrated in Figure \ref{fig:model_architecture}, we introduce CoMe-VLA, a cognitive and memory-aware VLA framework that leverages large-scale human egocentric priors with robotic execution capabilities for active perception. We detail its model architecture in Section \ref{sec_model_architecture} and training strategy in Section \ref{sec_training_strategy}.

\subsection{Model Architecture}
\label{sec_model_architecture}

We build CoMe-VLA upon Qwen3-VL-2B \cite{bai2025qwen3vltechnicalreport} in GR00T \cite{bjorck2025gr00t} style, to leverage strong visual understanding and semantic reasoning. CoMe-VLA takes as input the temporal (historical and current) egocentric visual observations, task descriptions, a cognitive token, and temporal proprioceptive states. These inputs are processed to predict an action chunk of $K$ future actions $\mathbf{A}_t \in \mathbb{R}^{K\times29}$. Each action vector within the chunk comprises the Cartesian position and 6D rotation representation \cite{rot6d} for the viewpoint and bimanual end-effectors, along with scalar bimanual gripper states. Concretely, CoMe-VLA realizes the information-seeking and decision-branching behaviors of active perception through learnable cognition and memory mechanisms, elaborated as follows.

\textbf{Cognitive Auxiliary Head.} We append a learnable cognitive token \texttt{[COG]} to the input sequence, aggregating visual and semantic histories. The token’s hidden state is then fed into a lightweight MLP-based cognitive auxiliary head to predict a binary label $c_t$, indicating whether enough task-relevant information has been obtained (e.g., the target is found and ready for manipulation). During inference, this label acts as a reliable monitor that informs the policy when to shift between exploratory and exploitative strategies. This transition is manifested through the switching of sub-task textual instructions, thus reconditioning the policy on a new goal once the information uncertainty in the preceding phase is successfully resolved.


\textbf{Dual-Track Memory.}
Active perception requires the agent to stay aware of what it has seen and done.  CoMe-VLA supports this through a dual-track memory system. Visual context is preserved by feeding a temporal window of egocentric observations into Qwen3-VL, while motor dynamics are captured by a transformer encoder operating on proprioceptive states. The action decoder cross-attends to both streams at each layer. Concretely, the memory window consists of the current frame and five historical frames sampled over the past 5 seconds. Although longer horizon or finer-grained temporal sampling are possible, this design is efficient and sufficient to support the challenging exploration in our active perception tasks, see Section \ref{sec_ablation} for experimental results.



\textbf{Implicit Data-Driven Decisions.}
A common pitfall in active perception models is the regression toward linear execution, where the policy produces the same behavior regardless of what is observed. CoMe-VLA avoids this by reasoning over history and current observation to implicitly capture both exploratory and exploitative decision paths mentioned in Section \ref{sec_core_mechanisms}. Training on large-scale human egocentric data, the model learns to resolve divergent perceptual outcomes, capturing branching behaviors as continuous variations in the perception-action coupling rather than pre-defined choices.

\subsection{Training Strategy}
\label{sec_training_strategy}

Our training pipeline consists of three progressive stages, designed to establish robust cognitive awareness first, followed by extracting general active perception and manipulation priors from large-scale human demonstrations, and finally grounding these capabilities in specific scenes for robot execution.

\textbf{Stage 1: Cognitive State Pretraining.} The objective of the first stage is to establish a foundational understanding of task progress from large-scale human data. Only the parameters of the vision-language model and the cognitive auxiliary head are updated in this stage, with all other modules frozen. Specifically, we use focal loss \cite{lin2017focal} for supervision:
\begin{equation}
    \mathcal{L}_1 = \mathcal{L}_{cog}= -\alpha_t(1 - p_t)^\gamma \log(p_t),
\end{equation}
where $p_t$ is the predicted probability of cognitive label, $\alpha_t$ and $\gamma$ are balancing and focusing factors, respectively. 

\textbf{Stage 2: Cognition-Action Joint Pretraining.} In the second stage, we unfreeze the rest of the modules to learn the active perception and manipulation priors in the human data. Alongside the cognitive focal loss, we introduce MSE loss \cite{wang2009mse} to supervise the action decoder's predicted velocity on five action components: viewpoint rotation ($\mathbf{v}_{vr}$) and position ($\mathbf{v}_{vp}$), bimanual end-effector rotation ($\mathbf{v}_{er}$) and position ($\mathbf{v}_{ep}$), bimanual gripper state ($\mathbf{v}_{g}$). The action loss is calculated as:
\begin{equation}
    \mathcal{L}_{action} = \sum_{i \in \{vr, vp, er, ep, g\}} \lambda_{i} \| \hat{\mathbf{v}}_i - \mathbf{v}_i^* \|_2^2,
\end{equation}
where $\hat{\mathbf{v}}_i$ and $\mathbf{v}_i^*$ are the predicted and ground-truth velocity sequences, respectively. The total loss is:
\begin{equation}
    \mathcal{L}_{joint} = \lambda_t \mathcal{L}_{cog} + \mathcal{L}_{action}.
    \label{eq:l_joint}
\end{equation}

\textbf{Stage 3: Robot Data Finetuning.} The final stage adapts the model to the robot's real dynamics and execution environments, while preserving the cognitive foundation, action and perception capabilities established in earlier stages. We switch to robot data and continue full-model optimization with the same loss structure as stage 2 defined in Eq. \eqref{eq:l_joint}.

%% file: sections/6_experiments.tex
\section{Experiments}

In this section, we conduct a series of experiments to evaluate the effectiveness of our proposed method, and answer the following questions:

\textbf{Q1: } How effectively does our proposed method perform on various long-horizon active perception tasks?

\textbf{Q2: } What is the impact of large-scale human egocentric pretraining to learn generalizable active perception priors?

\textbf{Q3: } Can cognition-based task decomposition and dual-track memory system enhance the model's reasoning capabilities?

\textbf{Q4: } Does our proposed method exhibit long-term robustness and dynamic adaptability when faced with uncertain environmental perturbations?

\subsection{Implementation Details}

\begin{figure*}[t]
    \centering
    \includegraphics[width=\linewidth]{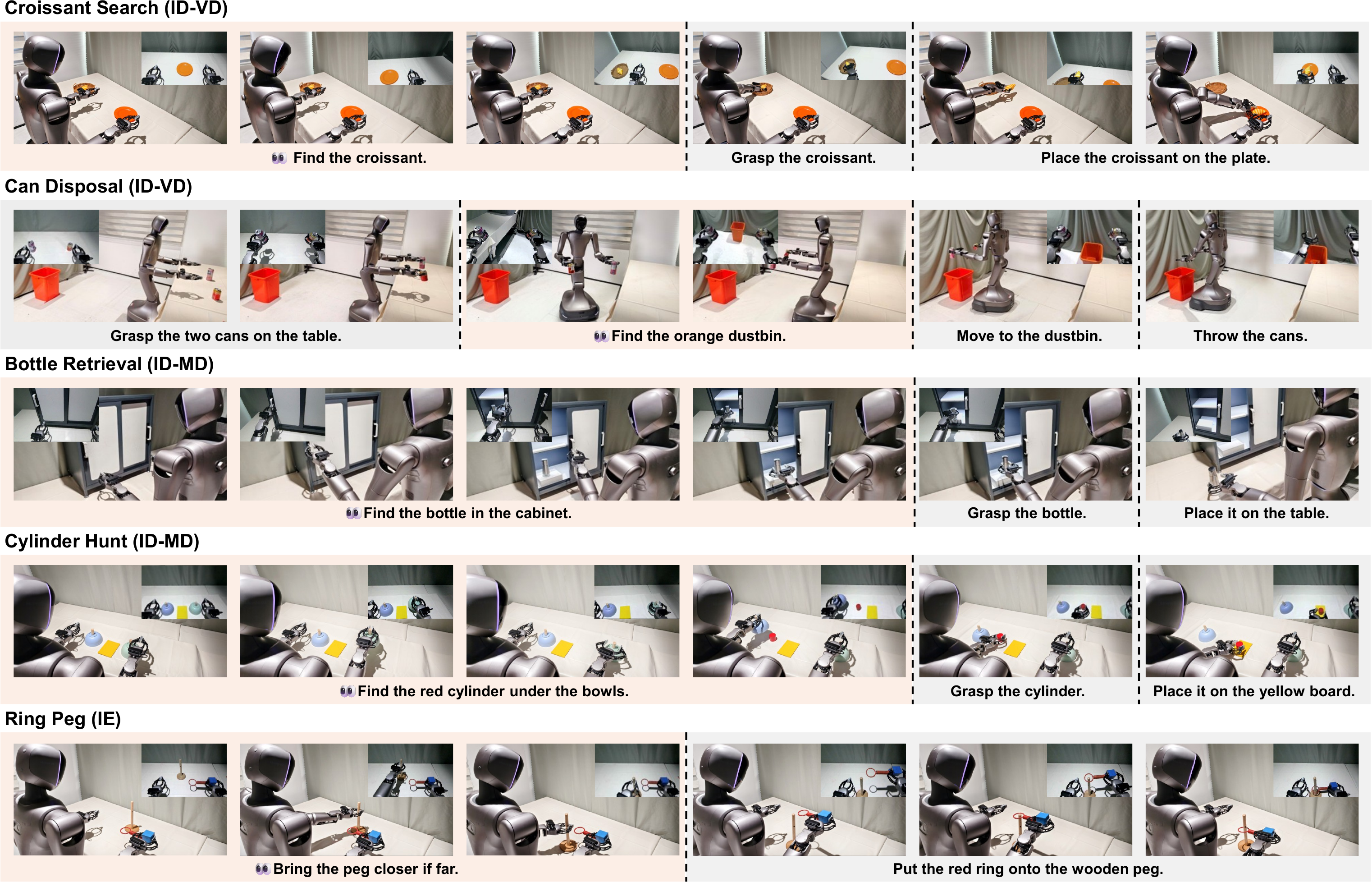}
    \setlength{\abovecaptionskip}{-10pt}
    \caption{\textbf{Evaluated Tasks.} All tasks are designed with uncertain initial conditions, where the locations of the target or task-critical objects are unknown to the model before execution, and can vary across multiple configurations. }
    \label{fig:tasks}
\end{figure*}

\subsubsection{Task Setup}
According to the paradigms discussed in Sec. \ref{sec_paradigm}, we consider 5 tasks in our experiments, as illustrated in Figure \ref{fig:tasks}.

\begin{itemize}[topsep=0pt]
    \item \textbf{Croissant Search (ID-VD):} The robot performs a head scan to locate a croissant that is initially out of the view, followed by a pick-and-place operation. The croissant may appear on the left or right side of the table.
    \item \textbf{Can Disposal (ID-VD):} The robot grasps two cans, moves its chassis to search for a dustbin initially out of view, and executes a release. The dustbin can be located in different unseen regions in the room.
    \item \textbf{Bottle Retrieval (ID-MD):} The robot opens a cabinet to reveal occluded shelves, identifying the bottle’s located tier to grab it to the table. The bottle may appear on either the upper or lower shelf.
    \item \textbf{Cylinder Hunt (ID-MD):} The robot uncovers two inverted bowls to locate a hidden cylinder, and places it at a goal location. The cylinder may be initially hidden under either the left or right bowl.
    \item \textbf{Ring Peg (IE):} The robot brings the peg closer if far, then inserts a ring onto it precisely. The peg initially appears either near or far from the robot.
\end{itemize}

\subsubsection{Training Setup}
We train CoMe-VLA with 800k human data samples (10 Hz, 22.2 hrs), and per-task 100k robot data samples (30 Hz, 0.9 hrs).  The entire 3-stage training is completed on 8 NVIDIA H100 HBM3 80G GPUs for 3 days. Please refer to the supplementary material for more details.


\subsection{Evaluation Protocols}
\subsubsection{Baselines} We compare our method with two types of baselines: (1) General-purpose VLAs, including OpenVLA-OFT \cite{kim2025openvla-oft} and $\pi_{0.5}$ \cite{black2025pi_0_5}; (2) Task-specific policies, including ACT \cite{act} and Diffusion Policy (DP) \cite{chi2025diffusion}. See baseline implementation details in the supplementary material.


\subsubsection{Metrics} We adopt the following two metrics for a comprehensive evaluation: (1) Success Rate (SR). The percentage of trials where the robot  completes the final stage of the task. (2) Search Time (ST). The average time from the search start until the target is localized and centered in view, averaged across trials per task, with failed trials capped at 1000 seconds. See supplementary material for per-task ST computation details.

\subsection{Comparative Evaluation}

\begin{table*}
    \centering
    \caption{Evaluation on 5 active perception tasks.}
    \resizebox{\textwidth}{!}{
    \begin{tabular}{c c c  c c  c c  c c  c c  c c}
        \toprule
            \multirow{2}{*}{Method} & \multicolumn{2}{c}{Croissant Search} & \multicolumn{2}{c}{Can Disposal} & \multicolumn{2}{c}{Bottle Retrieval}  & \multicolumn{2}{c}{Cylinder Hunt}  & \multicolumn{2}{c}{Ring Peg} & \multicolumn{2}{c}{Mean} \\  
            \cmidrule(lr){2-3} \cmidrule(lr){4-5}  \cmidrule(lr){6-7}  \cmidrule(lr){8-9}  \cmidrule(lr){10-11} \cmidrule(lr){12-13} 
              & SR $\uparrow$ & ST  $\downarrow$ & SR $\uparrow$ & ST $\downarrow$ & SR  $\uparrow$ & ST $\downarrow$ & SR $\uparrow$ & ST  $\downarrow$ & SR $\uparrow$ & ST  $\downarrow$ & SR $\uparrow$ & ST  $\downarrow$\\
            \midrule
            OpenVLA-OFT \cite{kim2025openvla-oft} & 3/30 & 540.7s & 7/30 & 760.8s & 1/30  & 843.6s & 2/30 & 938.0s & 6/30 & 617.3s &12.7\% & 740.1s\\
            $\pi_{0.5}$ \cite{black2025pi_0_5} & 5/30 & 116.6s & 6/30 & 867.5s & 4/30 &  757.1s & 2/30 & 837.2s & 7/30 & 501.4s &16.0\% & 616.0s \\
            ACT \cite{act} & 0/30 & 591.9s & 0/30 & 1000s & 0/30 & 1000s & 0/30 & 981.0s & 4/30  & 724.5s &  2.7\% & 859.8s \\
            DP \cite{chi2025diffusion} & 9/30 & 431.8s & 0/30 & 1000s & 3/30 & 766.4s & 5/30 & 785.8s & 8/30 & 559.5s & 16.7\% & 710.7s\\
            \midrule
            CoMe-VLA (0 + 400k)    & 6/30  & 237.8s & 15/30 & 458.5s & 10/30 & 403.1s & 13/30 & 383.3s & 20/30 & 162.5s & 42.7\% & 329.0s \\
            CoMe-VLA (400k + 400k) & 24/30 & 28.4s  & 18/30 & 158.1s & 16/30 & 336.2s & 22/30 & 107.5s & 28/30 & 91.4s & 72.0\% & 144.3s \\
            CoMe-VLA (800k + 400k) & 28/30 & 24.3s  & 24/30 & 138.3s & 21/30 & 157.4s & 28/30 & 88.2s  & 30/30 & 58.9s & 87.3\% & 93.4s  \\
            \hline 
            \rowcolor{yellow!30} \textbf{CoMe-VLA (800k + 100k)}\rule{0pt}{2.6ex} & 28/30 & 26.7s & 22/30 & 141.3s & 18/30 &168.0s & 27/30 & 91.0s & 30/30 & 62.6s  & 83.3\% & 97.9s \\
        \bottomrule
        \hline 
    \end{tabular}
    }
\label{tab_evaluation}
\end{table*}

We compare our proposed method with baselines and answer Q1. As shown in Table \ref{tab_evaluation}, our method consistently outperforms all baselines across different tasks. Specifically, our approach achieves a mean SR of 83.3\%, surpassing OpenVLA-OFT  (12.7\%), $\pi_{0.5}$ (16.0\%), ACT (2.7\%), and DP (16.7\%), while maintaining a favorable ST of 97.9s. Based on these results, we provide two interesting findings.

\textbf{Illusory Exploration.} Though some of the baselines are with valid ST values ($<$1000 seconds), this does not necessarily imply effective search capability. For instance, $\pi_{0.5}$ reports a competitive ST of 116.6s in the Croissant Search task. However, this performance actually stems from stochastic ``random-walk" behaviors rather than intentional exploration, often failing to stabilize when the target enters the field of view. Baselines like ACT and DP mostly exhibit huge action bias (e.g., consistently turning the head to the left), and their success often results from coincidental alignment between this bias and target locations. 

\textbf{Weak Visuo-Motor Grounding.} We observe that some baselines struggle with hand–eye coordination. For example, $\pi_{0.5}$ occasionally exhibits misaligned perception–action behaviors, such as gazing toward one direction while executing grasp actions in the opposite one, indicating insufficient grounding between visual attention and motor control.



\subsection{Data Composition Analysis}
 We answer Q2 by training CoMe-VLA with different data compositions, where these variants are annotated with human and per-task robot sample counts used for training: (1) CoMe-VLA (0 + 400k). (2) CoMe-VLA (400k + 400k). (3) CoMe-VLA (800k + 400k). (4) CoMe-VLA (800k + 100k).

As shown in Table \ref{tab_evaluation}, adding 400k human samples to 400k robot data boosts success rate from 42.7\% to 72.0\%, and scaling to 800k human data further improves it to  87.3\%. 
Notably, with sufficient human data, reducing robot data to just 100k samples per task causes only a minor drop to 83.3\%. This demonstrates that large-scale human data provides a dense foundation of exploratory prior, effectively lowering the sample complexity for transferring to robot embodiments.

\subsection{Ablation Studies}
\label{sec_ablation}

We ablate dual-track memory system and cognition-based task decomposition strategy to answer Q3.


\subsubsection{Ablation on Memory} 
\begin{figure}[t]
    \centering
    \setlength{\abovecaptionskip}{-10pt}
    \begin{minipage}{0.57\linewidth}
        \centering
        \includegraphics[width=\linewidth]{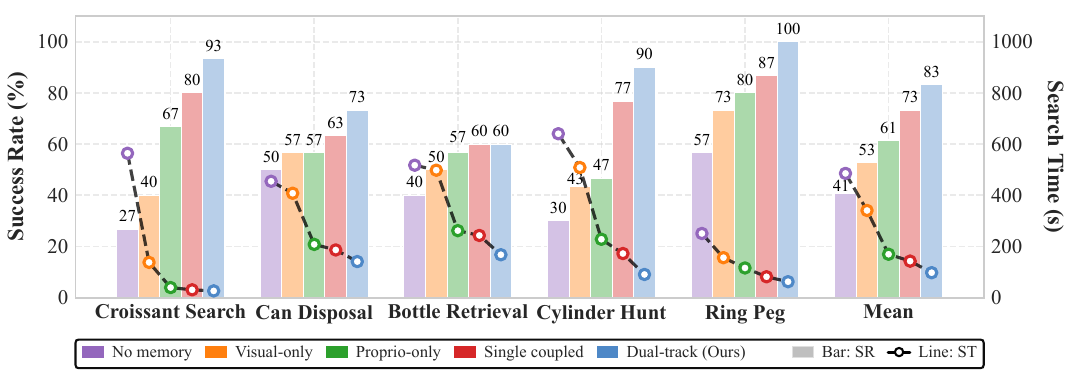}
        \vspace{16pt} 
        {\footnotesize (a) Memory Architecture}
        \label{fig:memory_design}
    \end{minipage}
    \hfill
    \begin{minipage}{0.41\linewidth}
        \centering
        \includegraphics[width=\linewidth]{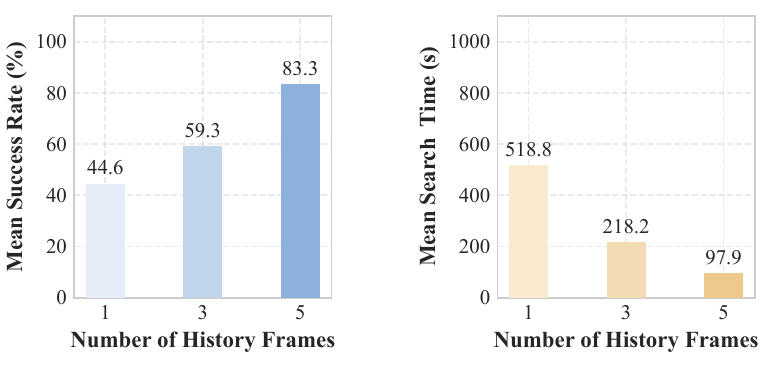}
        \vspace{16pt} 
        {\footnotesize (b) Memory Length}
        \label{fig:memory_length}
    \end{minipage}
    \caption{\textbf{Ablation on Memory Architecture and Temporal Horizon Design.}}
    \label{fig:memory}
\end{figure}

As shown in Figure \ref{fig:memory}, memory plays a critical role in active perception. Removing the memory drastically collapses task execution (dropping from 83.3\% to 40.7\% averaged over tasks),  confirming importance of modeling historical context for long-horizon active perception. We further analyze it from two perspectives as follows.


 \textbf{Memory Architecture.} 
 Replacing our dual-track memory with a single coupled memory fusion leads to clear performance drops, validating that decoupled visual–proprioceptive modeling better uses multimodal cues. In addition, we observe visual-only memory can sustain stable cognitive judgments but perform poorly in action execution.  In contrast, proprioception-only memory maintains better motion continuity but lack cognitive awareness and fail to adapt actions based on sensory feedback. These results confirm the necessity of dual-track memory for robust decision making and precise motor control.
 

\textbf{Memory Length.}  We additionally study the effect of memory temporal horizon, shown in the right of Figure \ref{fig:memory}. Our default setting of five historical frames substantially outperforms shorter histories (SR: 44.6\% $\rightarrow$ 83.3\%; ST: 518.8s $\rightarrow$ 97.9s), indicating that richer temporal context improves both perception and action quality. While longer histories could theoretically enhance performance, they introduced an increased computation burden due to longer visual token sequences within the VLM. The adopted memory configuration thus achieves a favorable balance between performance and computational feasibility for diverse active perception tasks. See the supplementary material for details.

\begin{wrapfigure}{r}{0.36\textwidth} 
    \centering
    \vspace{-10pt}  
    \subfigure[Success Rate]{
        \includegraphics[width=0.75\linewidth]{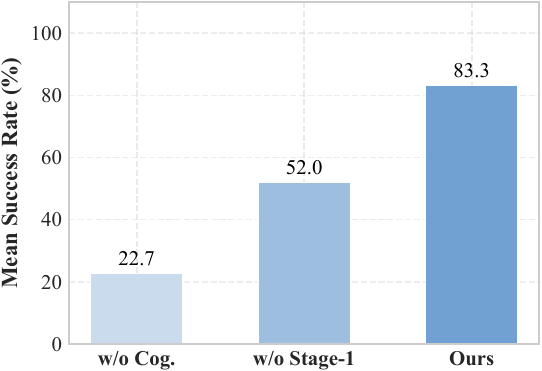}
        \label{fig:cognition_sr}
        }
        \subfigure[Search Time]{
            \includegraphics[width=0.75\linewidth]{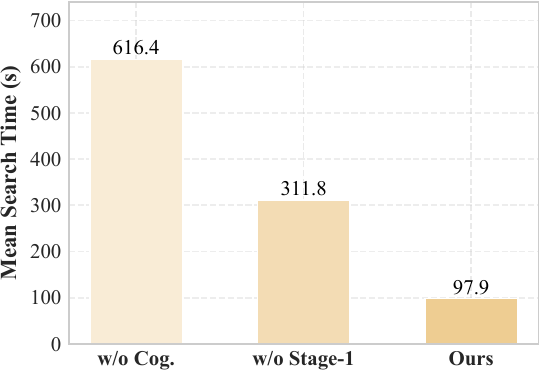}
            \label{fig:cognition_st}
            }  
        \caption{\textbf{Ablation on Cognitive Grounding.}}
        \label{fig:cognition}
    \vspace{-10pt} 
\end{wrapfigure}

\subsubsection{Ablation on Cognition}  As illustrated in Figure \ref{fig:cognition}, removing cognition-based task decomposition leads to a significant decline in success rates for long-horizon tasks. 
We attribute this to the tension between high-level semantic reasoning (e.g., assessing whether the target has been found) and low-level visuo-motor grounding required for precision control. Tightly coupling these two induces severe instability.

Removing cognitive pretraining (Stage 1) further hurts performance (SR drops 31.3\%, ST increases 213.9s), suggesting that jointly learning high-level task progress and low-level control from scratch impairs action-level and task-level feature extraction. In contrast, pre-learning cognitive representations enables the model to refine motor commands using high-level task awareness.

\subsection{Robustness Analysis}

To answer Q4, we evaluate the robustness of our framework under dynamic perturbations using a 10-minute adversarial test on the Croissant Search task. With grasping disabled to focus on active perception, we test two perturbations: (i) \textbf{Sudden Disappearance}: the target object is removed entirely from the workspace; and (ii) \textbf{Relocation}: the target is moved to an out-of-sight location. Results demonstrate that in the absence of the target, the robot exhibits persistent stochastic scanning without being trapped in a local fixed viewpoint. Once the target is reintroduced and enters the robot's field of view, the model rapidly achieves visual locking and maintains a sustained gaze. If the target disappears again, the robot reverts to scanning immediately. These seamless transitions between search and gaze modes demonstrate the framework’s high robustness in handling high uncertainty and dynamic environments.


%% file: sections/7_conclusion.tex
\section{Limitations and Future Work}

In this work, we formalize active perception as a history-dependent perception–action loop and propose CoMe-VLA, a cognitive and memory-aware framework that distills human-like exploration behaviors from large-scale egocentric data. By unifying the action space and adopting stage-wise training, our framework effectively transfers human priors to robotic systems. Extensive experiments validate the effectiveness of our cognition and memory mechanisms for robust, long-horizon active perception. Despite these advances, several challenges remain. Our fixed short memory window may not scale to extremely long horizons (e.g., tens of minutes) without introducing stale information, and embodiment-specific robot data is still required. Future work will explore adaptive memory structures and reinforcement learning to enable more autonomous and open-ended active perception.

%% file: sections/X_suppl.tex
\clearpage

\begin{center}
{\Large\bfseries Supplementary Material}
\vspace{2em}
\end{center}

\section{Robot Platform}

\subsection{Hardware}
\begin{figure}[H]
    \centering
    \includegraphics[width=0.5\linewidth]{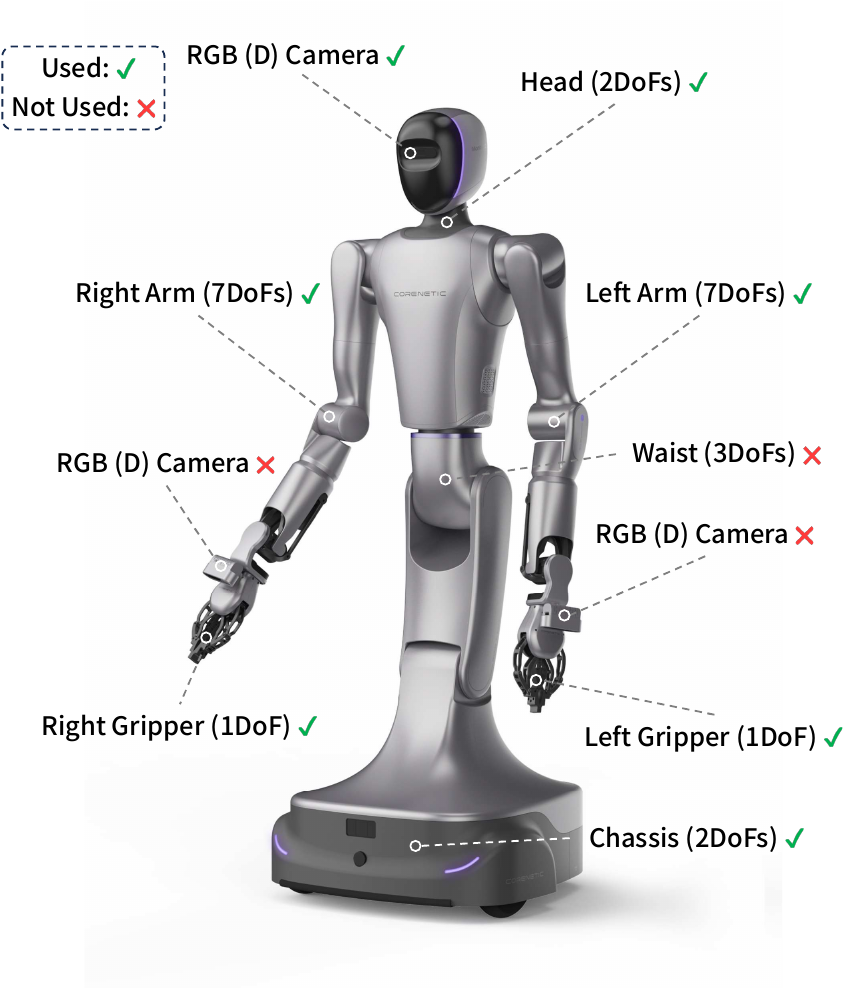}
    \caption{The Corenetic Monte02 robot platform.}
    \label{fig:robot_dofts}
\end{figure}

We conduct all experiments on the Corenetic Monte02 wheel-based humanoid robot \cite{corenetic}, as illustrated in Figure \ref{fig:robot_dofts}. The robot features 23 degrees of freedom (DoFs), including 2 DoFs for the mobile chassis (linear and angular velocities), 3 DoFs for the waist, 7 DoFs for each arm, 2 DoFs for the head, and 2 DoFs for the left and right grippers. It is equipped with three RGB-D cameras located at the head and both wrists. In our experimental setup, the waist joints are disabled, and only the RGB stream from the head-mounted camera is used as visual input, while the wrist cameras and depth channels are not involved in perception.

\subsection{Teleoperation}
\begin{figure}[H]
    \centering
    \includegraphics[width=0.7\linewidth]{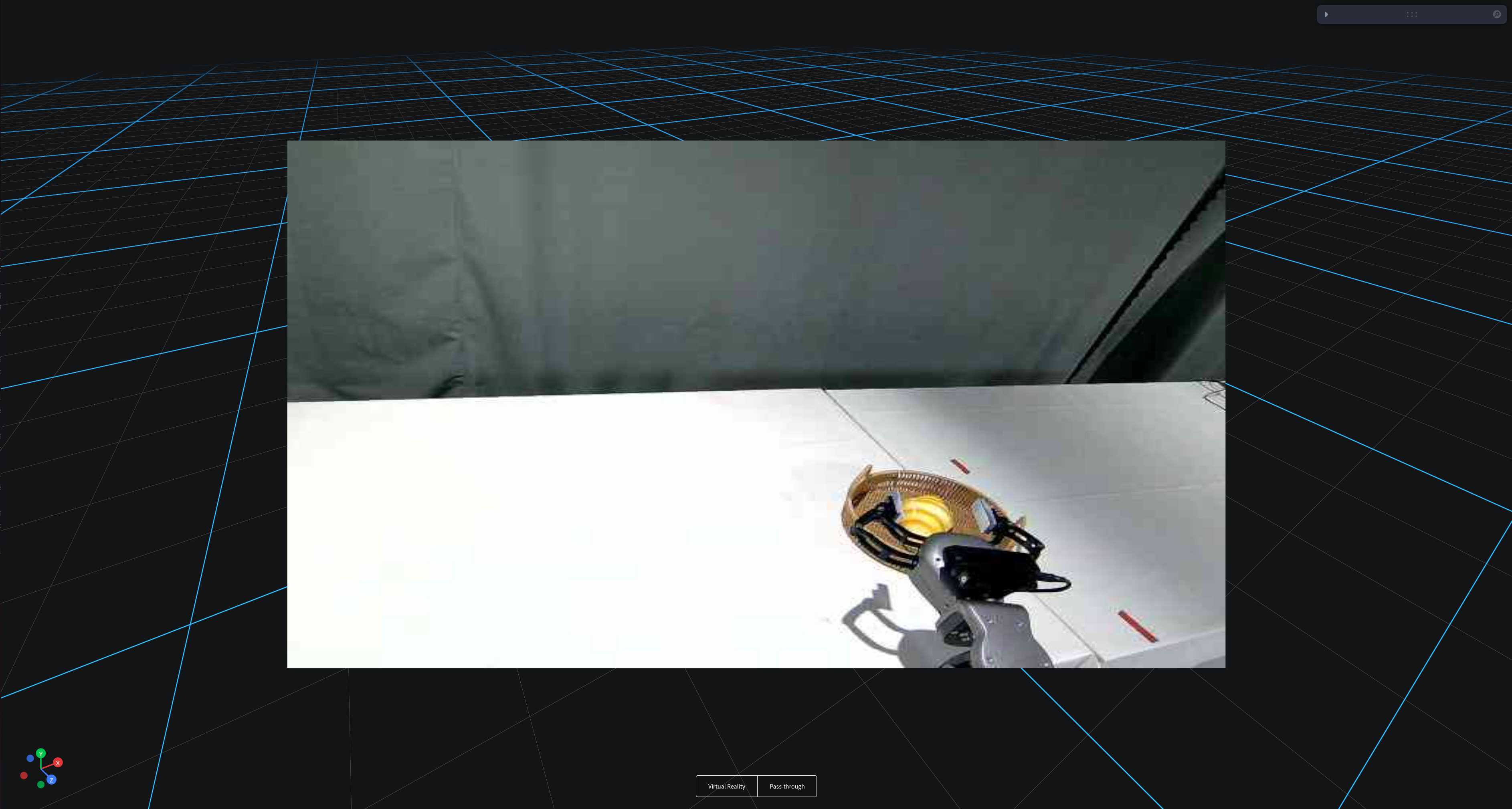}
    \caption{Interface of the VR-based teleoperation system.}
    \label{fig:vuer}
\end{figure}

Based on Vuer \cite{vuer}, we develop a VR-based teleoperation system, as illustrated in Figure \ref{fig:vuer}. Using a Meta Quest 3 headset \cite{meta}, the operator can control the robot’s head, dual arms, grippers, chassis, and waist through head tracking, controller tracking, and button inputs. To approximate an egocentric perception setting, we stream the robot’s head-mounted camera feed directly into the VR interface, allowing the operator to perform tasks from a near first-person perspective of the robot, which aligns with the requirements of active perception. During teleoperation, we record the robot’s head pose, dual-arm joint states, gripper widths, and chassis states, forming the robot-centric dataset used in our experiments.

The robot states and the egocentric head-mounted camera video streams are recorded at 30 Hz during teleoperation.

\section{Data Collection and Processing}

\subsection{Human Data Sources and Statistics}
We construct our human dataset from two large-scale, publicly available egocentric datasets, CaptainCook4D \cite{peddi2024captaincook4d} and Ego-Exo4D \cite{grauman2024ego}, selected according to the criteria described in the main paper, and their key statistics are summarized in Table~\ref{tab:human_data}. These datasets provide complementary coverage in terms of sensing fidelity, task structure, and behavioral diversity, which together support the learning of active perception policies. 
 
\begin{table*}
    \centering
    \caption{Statistics of Curated Human Data.}
    \label{tab:human_data}
    \resizebox{\textwidth}{!}{
    \begin{tabular}{l l c c c l}
    \toprule
         Dataset & Scenarios & Takes & Samples & Unique Lang. & Curation \& Filtering \\
    \midrule
    \textbf{CaptainCook4D} & Cooking & 257  & 1,071,604  & \ \ 349 & HoloLens GT \\
    \textbf{Ego-Exo4D} & Cook, Bike, Covid & 249 & \ \  421,582 & 2,730 & MANO optimization; strict visibility filtering   \\        
    \bottomrule
    \end{tabular}
    }
\end{table*}

\begin{table*}[t]
    \centering
    \caption{Robot Data Collection Statistics.} 
    \label{tab:robot_data_stats}
    
    \resizebox{\textwidth}{!}{
    \begin{tabular}{l c l l c }
    \toprule
    Task &  AP Type &  Randomization Config. & AP Protocol & Distribution \\
    \midrule
    
    \multirow{2}{*}{Croissant Search} & \multirow{2}{*}{ID-VD} & Target on \textbf{Left} & \multirow{2}{*}{\parbox{8.5cm}{Initial view obstructed; operator randomly scans left or right first.}} & 49.2\%  \\
     & & Target on \textbf{Right} & & 50.8\%  \\
    \cmidrule{1-5}
    
    \multirow{3}{*}{Can Disposal} & \multirow{2}{*}{ID-VD} & Bin \textbf{Behind} & \multirow{3}{*}{\parbox{8.5cm}{Rotate chassis clockwise to scan the environment; explicitly center the bin in the camera view before navigating.}} & 28.2\%  \\
     & & Bin \textbf{Right} & & 28.2\%   \\
     & & Bin \textbf{Right-Back} & & 43.6\%  \\
    \cmidrule{1-5}
    
    \multirow{2}{*}{Bottle Retrieval} & \multirow{2}{*}{ID-MD} &\textbf{Top} Shelf & \multirow{2}{*}{\parbox{8.5cm}{Target initially occluded; open cabinet to reveal target, then perform gaze fixation and hand alignment to the active shelf level.}} & 50.0\%  \\
     & & \textbf{Bottom} Shelf & & 50.0\%  \\
    \cmidrule{1-5}
    
    \multirow{2}{*}{Cylinder Hunt} & \multirow{2}{*}{ID-MD} &Under \textbf{Left} Bowl & \multirow{2}{*}{\parbox{8.5cm}{Systematic search: open right bowl first; stop if found, otherwise open left.}} & 50.0\%  \\
     & & Under \textbf{Right} Bowl & & 50.0\%  \\
    \cmidrule{1-5}
    
    \multirow{2}{*}{Ring Peg} & \multirow{2}{*}{IE} & Peg \textbf{Near} & \multirow{2}{*}{\parbox{8.5cm}{Adaptive strategy: Direct placement for near targets vs. Retrieve-then-place for far targets.}} & 49.2\%  \\
    & & Peg \textbf{Far} & & 50.8\%  \\
     
    \bottomrule
    \end{tabular}
    }
 
\end{table*}

\subsubsection{CaptionCook4D} CaptionCook4D is captured using a HoloLens device \cite{hololens} and provides high-quality egocentric visual observations together with accurate head and hand actions and manipulation annotations. The dataset primarily consists of structured cooking procedures with relatively fixed execution flows, while still allowing variations in low-level actions due to differences in performers and environments. The curated subset contains approximately 1.07M samples across 257 takes, covering 349 unique language labels.


\subsubsection{Ego-Exo4D} In contrast to the structured environment of CaptainCook4D, EgoExo4D offers data captured via Aria glasses \cite{aria} in diverse, unconstrained real-world settings. We target the Cooking, Bike Repair, and Covid subsets, as these scenarios inherently require active visual search in cluttered scenes. However, a major challenge with Ego-Exo4D is that the raw annotations provide wrist positions but lack the accurate 3D rotation required for our unified action space. To mitigate this, we discard frames with unreliable hand tracking, retaining only samples where each hand has more than 16 (out of 21) valid joints. We then apply MANO-based \cite{mano} optimization on these filtered samples to recover accurate 6DoF hand poses. After filtering, the resulting subset comprises $\sim$421k samples across 249 takes, spanning 2,744 language labels, which preserves diverse interaction patterns while ensuring consistent proprioceptive representations.

To construct our final training dataset, we randomly select 400k samples from each of the two curated datasets, which yields a unified 800k-sample human dataset for model pretraining.

\subsection{Robot Data Collection Protocol}
To ensure that the policy learns genuine active perception capabilities rather than memorizing spatial trajectories, we enforce a strict ``blind'' data collection protocol. An experimenter randomizes the scene layout (e.g., target locations) for each episode while the operator is visually isolated. Visual access to the robot's onboard camera feed via the VR interface is granted only upon episode initiation, necessitating immediate visual exploration to locate the task targets.

We design five distinct tasks to cover a spectrum of active perception challenges. For each task, operators follow corresponding behavioral protocols to ensure high-quality and consistent demonstrations. Table \ref{tab:robot_data_stats} details the specific randomization configurations and the active perception (AP) strategies employed for each task.

\subsection{Unified Data Representation and Processing}
\subsubsection{Head and Hand Poses}

All poses are expressed in a unified body-centric local frame $\mathcal{L}$, defined as a right-handed coordinate system shared across human and robot embodiments. The base frame is initialized using the head pose in the first frame of each episode, and all subsequent observations and actions are represented relative to this reference via rigid-body transformations in $\mathbf{SE}(3)$. 

\textbf{Human Data.} Human egocentric datasets provide head and wrist poses in a world frame $W$ that varies across episodes. At each timestep $t$, these are expressed as homogeneous transformation matrices:
\begin{equation}
    T_H^W(t),\ T_L^W(t),\ T_R^W(t) \in \mathbf{SE}(3),
\end{equation}
where
\begin{equation}
    T = \begin{bmatrix}
        R & p \\
        0 & 1
    \end{bmatrix}, \quad R \in \mathbf{SO}(3),\ p \in \mathbb{R}^3.
\end{equation}
Here,  $H$ denotes the head, $L$ and $R$ the left and right wrists, and $W$ the world frame.

We define the episode base frame $B$ as the initial head pose 
\begin{equation}
    T_B^W = T_H^W(0).
\end{equation}
Subsequent poses are then transformed into this base frame:
\begin{gather}
T_i^B(t) = \text{inv}(T_B^W) \cdot T_i^W(t) \cdot T_{i_{j} \to \mathcal{L}} , \notag \\
i \in \{H, L, R\}, \quad j \in \{\text{CaptainCook4D}, \text{Ego-Exo4D}\}.
\end{gather}

\textbf{Robot Data.} For the Corenetic Monte02 platform, raw data includes head Euler angles $[\theta_H(t), \psi_H(t)]$ (pitch and yaw) and chassis poses in the odomtry frame alongside end-effector poses relative to the chassis:
\begin{equation}
    T_C^O(t),\ T_L^C(t),\ T_R^C(t) \in \mathbf{SE}(3),
\end{equation}
with $C$ denoting the chassis base, $L$ and $R$ the left and right end-effectors, and $O$ the odometry frame.

\paragraph{Head-Chassis Aggregation and Decomposition} To mimic the human's integrated perception, we aggregate the chassis and head movements. The global head rotation is derived from the chassis yaw and the robot head angles,
\begin{equation}
R_H^O(t) = \text{Rot}_z(\psi_C(t) + \psi_H(t)) \cdot \text{Rot}_y(\theta_H(t)).
\label{eq: head_rot}
\end{equation}
And the composite head position becomes $p_H(t) =[x_C(t), y_C(t), z_{\text{fixed}}]^T$, where $z_{\text{fixed}}$ is the constant head height. Together with the rotation in Eq. \eqref{eq: head_rot}, this defines the head pose $T_H^O$. 


During inference, the target head pose is decomposed into chassis and head gimbal commands via a threshold-based heuristic. Specifically, if the target necessitates planar translation in $(x,y)$ or involves a significant angular deviation (e.g., $|\psi|>0.7\ \text{rad}$, the chassis takes primary control to realize the desired head pose. Conversely, for tabletop manipulation tasks that do not require base relocation, the chassis remains stationary, and the motion is executed solely by the head gimbal. This mechanism ensures that long-range motion tasks, such as \textit{Can Disposal}, are handled by the chassis, while fine-grained observation is managed by the head.

\paragraph{Frame Transformations} We define the episode base frame $B$ consistently as the initial head frame rigidly attached to the chassis. From the robot's URDF links, the static head-to-chassis transform is:
\begin{equation}
    T_H^C(0) = \begin{bmatrix}
        I & [0, 0, z_{\text{fixed}}]^T \\
        0 & 1
    \end{bmatrix},
\end{equation}
where $I$ is the identity rotation matrix. Thus, 
\begin{equation}
    T_B^O = T_H^O(0) =  T_C^O(0) \cdot T_H^C(0).
\end{equation}
The synchronized head and hand poses in the unified base frame $B$ are computed as follows:
\begin{gather}
    T_i^B(t) = \text{inv}(T_B^O)  \cdot T_i^O(t) \cdot T_{i_j \to \mathcal{L}}, \notag \\ 
    i \in \{L,R,H\}, \quad  j\in \{\text{Robot}\}.
\end{gather}
where 
\begin{gather}
T_i^O(t) =  T_C^O(t)  \cdot  T_i^C(t) , i \in \{L,R\}.
\end{gather}

\begin{table*}[t]
    \centering
    \caption{Settings of training hyperparameters in different training stages.}
    \resizebox{\textwidth}{!}{
    \begin{tabular}{c c c c c c c c c c c c c}
    \toprule
         Training Stage & Learning Rate & LR Scheduling &  Epoch & Action Chunk Size & $\alpha_t$ & $\gamma$ & $\lambda_t$ & $\lambda_{vr}$ & $\lambda_{vp}$ & $\lambda_{er}$ & $\lambda_{ep}$ & $\lambda_{g}$  \\
    \midrule
         Stage 1 & 2e-5 & Cosine & 10 & - & 0.25 & 2.0 & 1.0 & - & - & - & - & -  \\
         Stage 2 & 1e-4 & Cosine & 20 & 30 & 0.25 & 2.0 & 0.5 & 1.5 & 1.0 & 1.0 & 1.0 & 1.0  \\
         Stage 3 & 2e-5 & Cosine & 50 & 30 & 0.25 & 2.0 & 0.8 & 1.5 & 1.0 & 1.0 & 1.0 & 1.0  \\
    \bottomrule     
    \end{tabular}
    }
    \label{tab_training_parameters}
\end{table*}

\subsubsection{Gripper Actions}
Human data provides hand joint positions, from which we compute a scalar gripper aperture by averaging the Euclidean distances between the thumb tip and the tips of the remaining fingers. This value is then linearly normalized to the range $[0,1]$ to obtain a unified representation of hand opening. For robot data, the raw gripper width is similarly normalized to $[0,1]$ for consistency with the human-derived signal.

During inference, the policy outputs a normalized gripper value $a_{grip}$. To prevent actuation jitter and ensure stable grasping, we discretize this value using fixed thresholds:
\begin{equation}w_{t} =\begin{cases}
w_{max} & \text{if } a_{grip} > 0.7 \quad (\text{Open}) \\
0 & \text{if } a_{grip} < 0.3 \quad (\text{Close}) \\
w_{t-1} & \text{otherwise} \quad (\text{Hold})\end{cases}
\end{equation}
where $w_{\max} = 0.1\,\mathrm{m}$ is the maximum open width, and $w_{t-1}$ is the gripper command from the previous timestep.

\section{Model}

\subsection{Architecture}

CoMe-VLA is built upon Qwen3-VL-2B \cite{bai2025qwen3vltechnicalreport} (without the LM head) and contains approximately 2.79B parameters in total. Below, we detail the architecture of each component.

\textbf{Proprioceptive Memory Encoder.} The proprioceptive memory encoder consists of a one-layer MLP, followed by layer normalization and a two-layer transformer encoder. Each transformer layer uses 8-head self-attention with a hidden size of 2048 (which aligns with the last hidden state dimension of Qwen3-VL) and a feedforward dimension of 8192.
  
 \textbf{Cognitive Auxiliary Head.} Given that the vocabulary size of Qwen3-VL tokenizer is 151643, we select token 151621 as the cognitive input token, which is a rarely used one and does not overlap with any functional tokens (e.g., EOS). We observe that directly feeding the cognitive latent token into a MLP yields suboptimal performance. Instead, we design the cognitive token to query the hidden states of preceding tokens via a cross-attention module with 8 attention heads and a hidden dimension of 2048. The resulting attended representation is then passed through a lightweight two-layer MLP, which projects the hidden representation to a scalar cognitive score. During deployment, a threshold is applied to this score to determine sub-task termination and transition. The threshold value is provided in Section \ref{sec_label}.

\textbf{Flow-Matching Action Decoder.}  In the action decoder, two separate one-layer MLPs are used to project viewpoint action noise and manipulation action noise, respectively. We then concatenate the two noise representations into a unified sequence, while preserving their modality identities for self-attention masking. The combined sequence is processed by a transformer composed of six blocks. Each block contains one self-attention layer and two cross-attention layers, all configured with a hidden dimension of 2048 and 8 attention heads, followed by a feed-forward MLP with an intermediate dimension of 8192. Residual connections are applied after each attention operation. Two AdaRMSNorm \cite{adanorm} layers are inserted before and after the feed-forward MLP in each block to stabilize training.

We have also explored a variant of the action decoder, in which the self-attention mask of the manipulation noise was extended to partially attend to the viewpoint action noise, aiming to exploit predictive eye signals to enhance eye–hand coordination. However, empirical results indicate that this design yields neither performance improvement nor degradation. 

\subsection{Training}

\subsubsection{Data Annotation and Preprocessing}
To provide explicit supervision for the cognitive component of our framework, we manually annotate sub-task boundaries in each demonstration. For robot trajectories, the final 90 frames of each sub-task are designated as ``sub-task completion" and assigned a cognitive label $c_t=1$. For human data, this completion window is set to the final 30 frames.  All the preceding frames are assigned $c_t=0$. To maintain computational efficiency, all visual frames are resized to $224\times224$ before being fed into the VLM.


\subsubsection{Training Details}
We mix collected robot data from all tasks to train a unified CoMe-VLA model. Across all training stages, following the official guidance of Qwen3-VL, we only update the vision projector (visual merger), while keeping the language backbone, vision encoder, and most transformer blocks frozen to preserve pretrained visual and semantic knowledge.

We detail our training hyperparameters for the three stages in Table \ref{tab_training_parameters}.

\section{Deployment}
\label{sec_deployment}

\subsection{Inference and Action Execution}
We use a fixed number of denoising steps, set to 5 during flow-matching action inference. Empirically, increasing the number of denoising steps yields neither performance improvement nor degradation, and this configuration is therefore used consistently across all tasks. For real-time deployment, we adopt a receding-horizon control strategy. At each timestep $t$, the model predicts an action chunk of length $K = 30$. To balance inference speed and control smoothness, we execute the first $10$ steps of the predicted trajectory before re-planning. All predicted continuous actions are clipped to the robot's physical limits (e.g., joint positions and velocities) before execution. This procedure is repeated at every control cycle. 

\subsection{Cognitive Label and Sub-task Transitions}
\label{sec_label}
We rely on a learned cognitive label to trigger transitions between exploration and exploitation behaviors. To suppress spurious activations caused by noisy predictions, we apply a label threshold  $\tau=0.7$ and require the predicted label to remain above this value for at least $3$ consecutive timesteps before it is considered valid. For multi-stage tasks, we reset the memory module at sub-task boundaries.
Empirically, we observe no significant performance difference between retaining and clearing memory across sub-tasks, and thus adopt memory clearance in deployment for simplicity.

\subsection{Task Evaluation Metrics}
\textbf{Success Rate (SR)} quantifies the percentage of trials in which the robot completes the entire task, including all intermediate sub-tasks. All outcomes are validated through post-hoc analysis of execution logs and video recordings to ensure a reliable and reproducible assessment of active perception performance.

\textbf{Search Time (ST)} is defined as the duration of the active perception phase for each task. Rather than relying solely on the label, ST is measured according to task-specific criteria to ensure consistent evaluation across all experimental settings.
\begin{itemize}
    \item Croissant Search: from the beginning of the episode until the croissant is centered in the robot’s field of view.
    \item Can Disposal: from the moment the robot picks up the bottle until the target dustbin is centered in the robot’s field of view.
    \item Bottle Retrieval: from the start of the episode until the target bottle is visible and the robot’s hand is aligned with it.
    \item Cylinder Hunt: from the start of the episode until the target cylinder is fully exposed.
    \item Ring Peg: from the start of the episode until the peg is brought to a nearby position ready for insertion.
\end{itemize}

\begin{table*}[t]
    \centering
    \caption{Performance of Different Cognition Designs}
    \resizebox{\textwidth}{!}{
    \begin{tabular}{c c c  c c  c c  c c  c c  c c}
    \toprule
        \multirow{2}{*}{Method} & \multicolumn{2}{c}{Croissant Search} & \multicolumn{2}{c}{Can Disposal} & \multicolumn{2}{c}{Bottle Retrieval}  & \multicolumn{2}{c}{Cylinder Hunt}  & \multicolumn{2}{c}{Ring Peg} & \multicolumn{2}{c}{Mean} \\  
        \cmidrule(lr){2-3} \cmidrule(lr){4-5}  \cmidrule(lr){6-7}  \cmidrule(lr){8-9}  \cmidrule(lr){10-11} \cmidrule(lr){12-13} 
          & SR $\uparrow$ & ST  $\downarrow$ & SR $\uparrow$ & ST $\downarrow$ & SR  $\uparrow$ & ST $\downarrow$ & SR $\uparrow$ & ST  $\downarrow$ & SR $\uparrow$ & ST  $\downarrow$ & SR $\uparrow$ & ST  $\downarrow$\\
        \midrule
        w/o Cog. & 6/30 & 540.2s & 5/30 & 820.5s & 6/30  &  680.8s & 5/30 & 710.3s & 12/30 & 330.2s & 22.7\% & 616.4s\\
        w/o Stage-1  & 13/30 & 148.4s & 14/30 & 480.5s & 16/30 & 425.6s & 14/30 & 310.2s & 21/30 & 194.3s & 52.0\% & 311.8s \\
        \midrule
        \textbf{Ours} & 28/30 & 26.7s & 22/30 & 141.3s & 18/30 &168.0s & 27/30 & 91.0s & 30/30 & 62.6s  & 83.3\% & 97.9s \\
    \bottomrule
    \hline 
    \end{tabular}
    }
    \label{tab_cognition}
\end{table*}

\begin{table*}[t]
    \centering
    \caption{Performance of Different Memory Length Choices}
    \resizebox{\textwidth}{!}{
    \begin{tabular}{c c c  c c  c c  c c  c c  c c}
    \toprule
        \multirow{2}{*}{Method} & \multicolumn{2}{c}{Croissant Search} & \multicolumn{2}{c}{Can Disposal} & \multicolumn{2}{c}{Bottle Retrieval}  & \multicolumn{2}{c}{Cylinder Hunt}  & \multicolumn{2}{c}{Ring Peg} & \multicolumn{2}{c}{Mean} \\  
        \cmidrule(lr){2-3} \cmidrule(lr){4-5}  \cmidrule(lr){6-7}  \cmidrule(lr){8-9}  \cmidrule(lr){10-11} \cmidrule(lr){12-13} 
          & SR $\uparrow$ & ST  $\downarrow$ & SR $\uparrow$ & ST $\downarrow$ & SR  $\uparrow$ & ST $\downarrow$ & SR $\uparrow$ & ST  $\downarrow$ & SR $\uparrow$ & ST  $\downarrow$ & SR $\uparrow$ & ST  $\downarrow$\\
        \midrule
        1 History Frame & 8/30 & 675.2s & 14/30 & 557.3s & 14/30  &  502.3s & 11/30 & 654.8s & 20/30 & 204.4s & 44.6\% & 518.8s\\
        3 History Frames  & 16/30 & 104.0s & 19/30 & 251.3s & 15/30 & 394.1s & 16/30 & 197.9s & 23/30 & 143.7s & 59.3\% & 218.2s \\
        \midrule
        \textbf{5 History Frames (Ours)} & 28/30 & 26.7s & 22/30 & 141.3s & 18/30 &168.0s & 27/30 & 91.0s & 30/30 & 62.6s  & 83.3\% & 97.9s \\
    \bottomrule
    \hline 
    \end{tabular}
    }
    \label{tab_memory_length}
\end{table*}

\begin{table*}[t]
    \centering
    \caption{Performance of Different Memory Architecture Choices}
    \resizebox{\textwidth}{!}{
    \begin{tabular}{c c c  c c  c c  c c  c c  c c}
    \toprule
        \multirow{2}{*}{Method} & \multicolumn{2}{c}{Croissant Search} & \multicolumn{2}{c}{Can Disposal} & \multicolumn{2}{c}{Bottle Retrieval}  & \multicolumn{2}{c}{Cylinder Hunt}  & \multicolumn{2}{c}{Ring Peg} & \multicolumn{2}{c}{Mean} \\  
        \cmidrule(lr){2-3} \cmidrule(lr){4-5}  \cmidrule(lr){6-7}  \cmidrule(lr){8-9}  \cmidrule(lr){10-11} \cmidrule(lr){12-13} 
          & SR $\uparrow$ & ST  $\downarrow$ & SR $\uparrow$ & ST $\downarrow$ & SR  $\uparrow$ & ST $\downarrow$ & SR $\uparrow$ & ST  $\downarrow$ & SR $\uparrow$ & ST  $\downarrow$ & SR $\uparrow$ & ST  $\downarrow$\\
        \midrule
        No memory & 8/30 & 564.2s & 15/30 & 454.8s & 12/30  &  517.6s & 9/30 & 640.5s & 17/30 & 251.0s &40.7 \% & 485.6s\\
        Visual-only  & 12/30 &138.1s & 17/30 & 408.1s & 15/30 &  498.2s & 13/30 & 508.6s & 22/30 &156.5s &52.7\% &340.9s \\
        Proprio-only & 20/30 & 39.8 s & 17/30 &  208.4s & 17/30 & 262.1s & 14/30 &228.4s & 24/30  & 116.4s &  61.3\% &  170.0s \\
        Single coupled & 24/30 & 31.0s & 19/30 & 186.7s & 18/30 &243.3s & 23/30 & 172.7s & 26/30 & 82.0s & 73.3\% & 143.1s\\
        \midrule
        \textbf{Dual-track (Ours)} & 28/30 & 26.7s & 22/30 & 141.3s & 18/30 &168.0s & 27/30 & 91.0s & 30/30 & 62.6s  & 83.3\% & 97.9s \\
    \bottomrule
    \hline 
    \end{tabular}
    }
    \label{tab_memory_design}
\end{table*}

\subsection{Chassis Control}
We adopt a velocity-based control scheme to drive the chassis toward target poses in the odometry frame. The robot executes navigation behaviors by regulating linear and angular velocities, enabling smooth and continuous motion of the mobile base while tracking the desired chassis pose. To ensure precise trajectory execution and safety, we implement a sequential proportional control strategy that decouples rotational and translational movements.

Given a decomposed target pose $\mathbf{p}_{tgt} = (x_t, y_t, \theta_t)$ and the current robot pose $\mathbf{p}_{curr} = (x, y, \theta)$, we define the distance error $e_d = \|(x_t, y_t) - (x, y)\|_2$, the heading alignment error $e_\psi = \operatorname{atan2}(y_t - y, x_t - x) - \theta$, and the final orientation error $e_\theta = \theta_t - \theta$. The controller operates as a three-stage state machine:
\begin{enumerate}
    \item Heading Alignment: The robot rotates in place to align its longitudinal axis with the bearing line to the target. To minimize rotation time, the controller is bi-directional: it selects the orientation (forward or backward) that requires the smallest angular displacement. The angular velocity is computed as $\omega = k_\omega e_\psi$.
    \item Linear Approach: Once aligned, the robot translates towards the target with linear velocity $v = k_v e_d$, while maintaining zero angular velocity.
    \item Final Orientation: Upon reaching the target position, the robot performs a final rotation to match the target yaw: $\omega = k_\omega e_\theta$.
\end{enumerate}

All velocity commands are clamped within predefined safety bounds, i.e., $v \in [v_{\min}, v_{\max}]$ and $\omega \in [\omega_{\min}, \omega_{\max}]$. To suppress oscillations near the target, dead-zone thresholds $\epsilon_p$ and $\epsilon_\theta$ are applied to translational and rotational errors, respectively. In our implementation, we use proportional gains of $k_v = 0.2$ and $k_\omega = 0.5$. The safety parameters are set to $v_{\max}=0.15\,\mathrm{m/s}$, $v_{\min}=0.08\,\mathrm{m/s}$, $\omega_{\max}=0.3\,\mathrm{rad/s}$, $\omega_{\min}=0.15\,\mathrm{rad/s}$, $\epsilon_p=0.015\,\mathrm{m}$,  $\epsilon_\psi=0.5\,\mathrm{rad}$ and $\epsilon_\theta=0.05\,\mathrm{rad}$. These parameters are empirically chosen and kept fixed across all tasks.

\subsection{Baseline}

We compare our method with two VLA-based baselines (OpenVLA-OFT \cite{kim2025openvla-oft} and $\pi_{0.5}$ \cite{black2025pi_0_5}) and two imitation learning baselines (ACT \cite{act} and Diffusion Policy \cite{chi2025diffusion}). It's worth mentioning that these baselines follow learning paradigms that are fundamentally different from ours, as they are optimized to learn from homogeneous robot-centric data. Unlike our framework, they lack the structural isomorphism mechanisms necessary to ingest raw human egocentric signals. To ensure a fair comparison and avoid introducing embodiment-specific noise into the baselines, we finetune or train these baselines exclusively on our robot-centric dataset, without incorporating human egocentric data.

Specifically, we finetune or train all the baselines using per-task 400k robot data samples. Aside from minor adjustments to the batch size, all other hyperparameters are kept consistent with their original implementations. We only expand the action output dimensions to accommodate the Corenetic Monte02 robot, without introducing any additional architectural modifications. The implementation details of each baseline are elaborated below. 

\subsubsection{VLA-based baselines}

For VLA-based baselines, which support text conditioning, we train each baseline with a single model on all collected robot data across tasks. Training follows our sub-task decomposition setting, where each sub-task is provided as a textual prompt during training and inference, enabling the models to condition their action generation on the current sub-task description.

Since these models do not explicitly provide cognitive signals for sub-task termination, we detect sub-task boundaries based on action consistency. 
Specifically, we trigger a transition when the average difference between consecutive actions over a 20-timestep sliding window falls below a threshold of $0.15$, serving as a proxy for sub-task policy convergence.

\subsubsection{Imitation learning baselines}

For imitation learning baselines, which are originally designed for single-task settings without text conditioning, we train a separate model for each task using full long-horizon trajectories. These models are thus required to directly learn end-to-end execution of the entire task sequence without any sub-task switching mechanism.

\section{Ablation and Additional Discussions}

\subsection{Ablation}

\subsubsection{Ablation on Cognition}
We conduct ablations on the cognition-based task decomposition strategy by replacing sub-task instructions with the full task instruction and disabling the task-switching mechanism during deployment. In addition, we ablate the cognitive state pretraining (Stage 1) by training the model using only Stage 2 and Stage 3, i.e., full-model training on human data followed by full-model fine-tuning on robot data, while keeping all other hyperparameters identical to those in the full setting. We detail the results of the ablations on cognition design in Table \ref{tab_cognition}.

\subsubsection{Ablation on Memory}
We detail the performance of different memory length choices in Table \ref{tab_memory_length}. Below we describe the implementation details of the ablations on the memory architecture.

\begin{itemize}
    \item No memory. We feed only the current visual frame and the current robot state into the model during training and deployment, without incorporating any historical information.
    \item Visual-only memory. We provide the model with the current and historical visual frames together with the current robot state, while excluding any proprioceptive history.
    \item Proprio-only memory. We incorporate only the current and historical proprioceptive states, together with the current visual frame, without using any visual history.
    \item Single coupled memory. We replace the two cross-attention layers in the action decoder with a single cross-attention layer that operates on a fused representation of visual and proprioceptive memory. Specifically, we fuse the VLM’s last hidden states and the temporal proprioceptive context by applying two cross-attention operations in a mutual querying manner. The resulting representations are then layer-normalized, concatenated, and passed through an MLP for fusion before being fed into the action decoder.
\end{itemize}

We detail the performance of these memory architecture choices in Table \ref{tab_memory_design}.

\subsection{Discussions of Related Active Perception Systems}
We acknowledge recent concurrent works such as ViA \cite{ViA}, ActiveUMI \cite{activeumi}, and EgoMI \cite{egomi}, which have made significant contributions to the field. Here, we clarify the differences in scope that guides our baseline selection. Works like ViA and ActiveUMI primarily focus on innovating teleoperation interfaces and data collection frameworks to enable the acquisition of high-quality active perception data. From an algorithmic perspective, these systems typically leverage established policy backbones to learn from the collected data. Since our experiments already include rigorous comparisons against these state-of-the-art policy architectures (e.g., Diffusion Policy \cite{chi2025diffusion}, $\pi_{0.5}$ \cite{black2024pi_0}), we consider the algorithmic performance aspect to be effectively covered without reproducing their specific hardware setups.
While EgoMI introduces algorithmic components such as a memory mechanism (SPARKS), this module is specifically designed to mitigate rapid viewpoint shifts and context loss inherent to their specific egocentric human demonstration pipeline.
In contrast, our method adopts a dual-track memory system and cognitive labels to formalize active perception as a non-Markovian decision process and autonomously trigger sub-task transitions. Given that these approaches emphasize different aspects of active perception and use distinct evaluation settings, direct quantitative comparison is impractical. Instead, we benchmark our proposed method against general-purpose imitation learning and VLA baselines to evaluate its performance.